\documentclass[10pt,twocolumn,letterpaper]{article}

\usepackage{wacv}              

\usepackage{graphicx}
\usepackage{amsmath}
\usepackage{amssymb}
\usepackage{booktabs}
\usepackage{xcolor}

\usepackage[pagebackref,breaklinks,colorlinks]{hyperref}

\usepackage[capitalize]{cleveref}
\crefname{section}{Sec.}{Secs.}
\Crefname{section}{Section}{Sections}
\Crefname{table}{Table}{Tables}
\crefname{table}{Tab.}{Tabs.}

\def\wacvPaperID{2} 
\def\confName{WACV}
\def\confYear{2025}

\begin{document}

\title{Experimenting with Affective Computing Models in Video Interviews with Spanish-speaking Older Adults}

\author{Josep López Camuñas\textsuperscript{1} \quad 
        Cristina Bustos\textsuperscript{1}\quad
        Yanjun Zhu\textsuperscript{2} \quad
        Raquel Ros\textsuperscript{3} \quad
        Agata Lapedriza\textsuperscript{1,2}\\
\textsuperscript{1}Universitat Oberta de Catalunya \quad \textsuperscript{2}Northeastern University \quad
\textsuperscript{3}PAL Robotics\\
\small {\tt \{jlopezcamu, mbustosro, alapedriza\}@uoc.edu},\\
\small {\tt ya.zhu@northeastern.edu,  raquel.ros@pal-robotics.com}
}

\maketitle

\begin{abstract}
Understanding emotional signals in older adults is crucial for designing virtual assistants that support their well-being. However, existing affective computing models often face significant limitations: (1) limited availability of datasets representing older adults, especially in non-English-speaking populations, and (2) poor generalization of models trained on younger or homogeneous demographics. To address these gaps, this study evaluates state-of-the-art affective computing models—including facial expression recognition, text sentiment analysis, and smile detection—using videos of older adults interacting with either a person or a virtual avatar. As part of this effort, we introduce a novel dataset featuring Spanish-speaking older adults engaged in human-to-human video interviews. Through three comprehensive analyses, we investigate (1) the alignment between human-annotated labels and automatic model outputs, (2) the relationships between model outputs across different modalities, and (3) individual variations in emotional signals. Using both the Wizard of Oz (WoZ) dataset and our newly collected dataset, we uncover limited agreement between human annotations and model predictions, weak consistency across modalities, and significant variability among individuals. These findings highlight the shortcomings of generalized emotion perception models and emphasize the need of incorporating personal variability and cultural nuances into future systems.

\end{abstract}

\section{Introduction}

\begin{figure*}[htbp]
\centering
\includegraphics[width=0.85\linewidth]{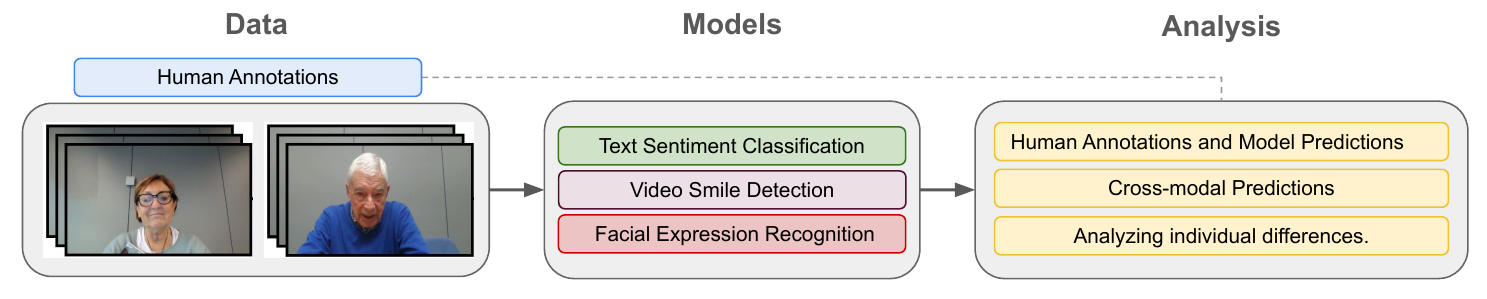} 
\caption{Overview of the analysis pipeline. Video and human annotations are processed through state-of-the-art machine learning models for text sentiment classification, smile detection and facial expression recognition. The outputs are analyzed in different studies to uncover key patterns and insights across multiple signals}
\label{fig:overview}
\end{figure*}

There is growing interest in developing virtual assistants and social assistive robots for older adults, as these technologies can encourage social interaction and promote well-being \cite{abdollahi2022artificial,abdollahi2017pilot,abdi2018scoping,brinkschulte2018empathic,justo2020analysis,palmero2023exploring}. For these systems to effectively engage older adults, they must be capable of perceiving emotional signals and personalizing interactions to accommodate the unique needs of this demographic \cite{panico2020ethical,troncone2020advanced}.

Despite advancements in emotion perception technologies for human-robot interaction \cite{mohammed2020survey,spezialetti2020emotion}, there remains a significant gap in research focused on older adults. This demographic is severely underrepresented in datasets used to train and evaluate affective computing models \cite{verhoef2023towards}, particularly older adults from non-English-speaking populations. As a result, current models often fail to generalize to elderly, leaving their emotional signals poorly understood or inaccurately interpreted. Addressing this gap is crucial, as older adults exhibit unique emotional and behavioral patterns influenced by age-related changes and cultural differences.

In this study, we investigate the performance of state-of-the-art facial emotion, text sentiment and smile recognition systems when applied to older Spanish-speaking adults, a group underrepresented in both affective computing research and publicly available datasets. To this end, we analyze emotional signals using two datasets: the Wizard of Oz (WoZ) dataset \cite{justo2020analysis}, which contains videos of older adults interacting with a virtual avatar, and a novel dataset we collected featuring human-to-human interviews with the same demographic. By introducing this new dataset, we provide a more diverse data distribution to enable broader and more generalizable insights into the emotional expressions of older Spanish-speaking adults.

Our work focuses on three key areas of analysis: the alignment between human-annotated labels and automatic model outputs, the relationship between facial emotion signals and speech sentiment, and the impact of individual differences in facial movements and sentiment expression. For this purpose, we leverage state-of-the-art, off-the-shelf models for facial expression recognition, smile detection, and text sentiment classification. An overview of our methodology is presented in Fig. \ref{fig:overview}, with a detailed description provided in Sec. \ref{sec:methods}.

Our findings reveal several challenges in applying generalized emotion perception models to this demographic. Specifically, we observed limited agreement between human annotations and model predictions and significant variability across individuals in facial movements and sentiment expression. These results highlight the limitations of current affective computing models in capturing the nuanced emotional signals of older Spanish-speaking adults and underscore the importance of incorporating personal and cultural variability in model design. By addressing these critical gaps, our work provides new insights into the development of inclusive affective computing systems. Upon acceptance, we will release all code to support further research and advancements in emotion perception for underrepresented populations.

\subsection{Small Data Statement}




This work focuses on analyzing the performance and output alignment of state-of-the-art affective computing models when applied to older adults, with a particular emphasis on Spanish-speaking populations. These populations are notably underrepresented in the development of machine learning models, which has resulted in systems that often fail to capture their unique emotional and behavioral patterns effectively. 
Studying this group is important because older adults have distinct needs and behaviors that differ significantly from younger or general adult populations. Factors such as age-related changes in facial expressions, speech patterns, and emotional dynamics necessitate tailored approaches for accurate emotion perception. Furthermore, Spanish-speaking older adults represent a significant portion of the global population, yet their cultural and linguistic nuances are seldom reflected in current datasets or models, which tend to focus on English-speaking demographics. 
Ensuring that affective computing models work well across all demographic groups, including underrepresented groups like the one studied in this paper, is critical for equitable technological advancements. If these systems are to support human well-being, they must accurately understand and respond to the emotional signals of diverse users. Failing to address these gaps risks perpetuating bias and excluding vulnerable populations from the benefits of such technologies. By focusing on this underrepresented group, the study contributes to developing more inclusive, robust, and effective affective computing systems that cater to the diverse needs of a global user base.

The goal of this work is to evaluate the performance of state-of-the-art affective computing models on Spanish-speaking older adults, an underrepresented demographic in existing datasets. To address this gap, we relied on the WoZ dataset and, complemented it by collecting a novel dataset to enable broader and more diverse analysis.  
Our findings reveal limited alignment between human-annotated emotional labels and model outputs, weak agreement between outputs from different modalities, as well as significant variability in facial movements and text sentiment expression across individuals. These results highlight the limitations of generalized models in capturing age- and culture-specific nuances, underscoring the need for inclusive approaches that account for such variability. By identifying these shortcomings, this study contributes to developing robust and equitable emotion perception systems that better serve diverse populations.

\section{Related Work}
\label{sec:related-work}

Interactive technologies, such as virtual agents and social robots, have demonstrated significant benefits in promoting older adults' health and social connection \cite{zhao2023technology}. These technologies feature non-intrusive, emotionally expressive virtual coaches or social robots that encourage users to adopt healthier lifestyles by improving nutritional habits, promoting physical activity, and facilitating social interactions. By fostering these behaviors, they help seniors reduce the risk of chronic diseases, enabling independent and fulfilling lives while also supporting their caregivers \cite{brinkschulte2018empathic}. Augmented and virtual reality technologies have also been explored to enrich older adults' experiences, including enhancing digital learning \cite{jin2024empowering}, improving long-term health interventions \cite{li2024designing}, and supporting overall well-being \cite{waycott2022role}. Social robots, in particular, have shown promise in providing high-quality care and social support. They have been studied for understanding their roles \cite{abdi2018scoping}, exploring stakeholder experiences with these systems \cite{abbot2019survey}, and demonstrating their effectiveness in improving older adults' well-being \cite{lee2022}. 

Research has investigated the generation of empathic behaviors in social robots using estimated affective states \cite{Paiva2017,Obaid2018,ElaheBagheri2020}. However, these studies primarily focus on general populations and do not specifically address elderly individuals. Recent work has emphasized the need for actively measuring engagement to drive human-robot interactions. For example, Zhang et al. \cite{Zhang2022} proposed a supervised machine learning approach to estimate the engagement states of older adults in multiparty human-robot interactions. Their method leverages pre-trained models to extract behavioral, affective, and visual signals, achieving effective engagement estimation by incorporating inputs from participants within the interaction group. Specific robots tailored for elderly populations include the DarumaTO robot \cite{du2023composite}, designed for Japanese-speaking users. It avoids camera use for privacy, relying instead on Speech Emotion Recognition (SER) through CNN and LSTM models to generate suitable facial responses. Similarly, Ryan \cite{abdollahi2022artificial}, a socially assistive robot for individuals with depression and dementia, uses multimodal emotion perception algorithms, including Residual Neural Networks (ResNet) for Facial Expression Recognition (FER) and natural language processing for SER. While empathic robots like Ryan have been perceived as more engaging and likable, the reported improvements were not statistically significant.

Recent advancements in deep learning methods have enabled multimodal approaches to emotion perception for older adults. Warnants et al. \cite{warnants2023implementing} introduced a digital platform for health monitoring, companionship, and emotional support. This platform combines SER with CNN-LSTM architectures to facilitate personalized care, highlighting the potential of machine learning to enhance emotional understanding and interaction in elderly populations. Despite these advances, datasets specifically targeting emotion estimation in older adults remain limited. ElderReact \cite{ma2019elderreact} is a multimodal dataset featuring over 1K annotated video clips of emotional responses from 40 aging participants. It includes visual and audio features, enabling analysis of which modality is more relevant for specific emotions. Unfortunately, this dataset was not accessible for our study. The WoZ dataset \cite{justo2020analysis}, part of the EMPATHIC project \cite{brinkschulte2018empathic}, offers a unique resource by providing interview-style video clips of older adults interacting with a simulated virtual agent. Palmero et al. \cite{palmero2023exploring} leveraged this dataset to explore non-verbal emotion expression recognition, applying deep learning models to integrate facial expressions, speech, gaze, and head movements. Their work identified emotional states such as puzzled, calm, and pleased. While previous research has advanced emotion perception in elderly, most studies focus on generalized models and fail to address individual variability or cultural nuances. Our work addresses these gaps by analyzing the alignment between human annotations and model outputs, exploring multimodal relationships, and assessing individual differences.

\section{Data}
\label{sec:data}
This research relies on two datasets focused on understanding emotional signaling in older adults, particularly among Spanish-speaking individuals. The first is the publicly available WoZ dataset, which examines interactions with virtual coaches. The second is the \emph{Short Interviews} dataset, a resource we developed to provide insights into natural, human-to-human interactions. 

\subsection{WoZ Dataset}
The WoZ (Wizard of Oz) dataset \cite{justo2020analysis} is designed to study how older adults interact with virtual coaches in the context of healthy aging. The data collection includes audio and video recordings, alongside dialogue transcriptions from sessions where seniors interacted with a simulated virtual coach. The dataset includes multimodal data—audio, video, and text transcriptions—annotated separately for emotional states across each modality.

For this study, we focused on a subset of 78 Spanish-speaking participants, consisting of 24 men and 54 women, all aged 65 and above. These individuals were selected from a larger group of 153 recruited through different community centers. Each participant took part in two structured sessions based on the GROW coaching model \cite{montenegro2019dialogue}, discussing topics such as leisure and nutrition. These sessions produced 4,529 audio-video segments annotated in different modalities like audio, text transcription and facial expressions. 
In this work, we use the emotional labels of speech transcription (\textit{positive, neutral} and \textit{negative}) and facial expressions (\textit{neutral, happy, pensive} and \textit{surprise}).
Neutral facial expressions were the most common, which aligns with the natural flow of conversations in this context. Annotations of facial expressions in video were carried out by two independent annotators, and any disagreements were resolved collaboratively to ensure reliable labeling. Further details on the data collection procedure and label distribution can be found in \cite{justo2020analysis, palmero2023exploring}.


\subsection{Our \emph{Short Interviews} Dataset}

The \emph{Short Interviews} dataset was created to feature older adults engaging in natural, human-to-human conversations. It captures semi-structured interviews between participants and a human interviewer, providing a more organic setting for studying emotional responses. This dataset includes 16 participants (13 women and 3 men) aged 62 and older (average age 77.3 years, standard deviation 7.9). Recruitment took place at a healthcare center offering group sessions focused on reducing social isolation among the elderly. Participants joined the study voluntarily after receiving detailed information about its purpose and procedures. Written informed consent was obtained from all participants, following ethical guidelines.

The interviews were conducted in a private, quiet room to ensure both comfort and high-quality recordings. Two cameras were used: one positioned on a table to capture a frontal view (recording at 1080p@30fps), and another placed laterally on a tripod (recording at 2160p@30fps). Audio was recorded with a cardioid microphone placed close to the participant to ensure clarity. The discussions covered several neutral topics, including food, music, and books, which were chosen to engage participants and elicit a broad range of emotional responses. This approach resulted in 3.5 hours of video, divided into 107 clips, 43 of which were focused on eliciting mild emotional reactions. While manual annotations have not yet been completed, the dataset provides a solid basis for developing emotion perception models.
In contrast to the WoZ dataset—which provides structured, annotated data centered on interactions with a virtual coach—the \emph{Short Interviews} dataset captures the spontaneity of human-to-human interactions, offering a more natural view of emotional responses. By combining these resources, we can explore both controlled and organic emotional dynamics.


\section{Methods}
\label{sec:methods}

Our analysis focuses on both verbal (speech transcriptions) and non-verbal (facial expressions and smile) behavioral signals in dyadic interactions. We use front-facing camera recordings that provide detailed views of participants' faces, along with their speech transcriptions. We employ three main components: smile detection, facial expression recognition, and text sentiment analysis. Additionally, we describe our methodology for aligning these different modalities temporally to enable cross-modal analysis.

\textbf{Smile Detection.} Smiles are one of the most significant affective signals in human interactions \cite{Eyben2011String-based, Johnston2010Why, MARTIN2017864, Mehu2007Smiles}, serving as key indicators of emotional states and social engagement. To detect smiles in video sequences, we finetune a Transformer-based video architecture using MARLIN \cite{cai2023marlinmaskedautoencoderfacial} as the backbone. The model takes a sequence of facial crops as input, extracted using the face detector from \cite{guo2020towards, 3ddfa_cleardusk}. We trained and tested the model on the CelebV-HQ dataset \cite{zhu2022celebvhq}, achieving 82\% accuracy (F1: 82\%) on training and 77\% accuracy (F1: 73\%) on test sets. The model processes 16-frame sequences and outputs a smile intensity score between 0 and 1. Qualitative evaluation using live webcam sequences highlighted the need to distinguish smiles from speech-related mouth movements. We set the detection threshold to 0.85 based on qualitatively analyzing smile intensities in these sequences, enabling robust separation of smiles from speech articulation while capturing both brief and prolonged smile expressions.


\textbf{Facial Expression Recognition.} For broader emotion analysis, we employ EfficientNet\cite{tan2020efficientnet} and MobileNet\cite{howard2017mobilenets} models trained on AffectNet\cite{mollahosseini2019affectnet}, following Savchenko \textit{et al}. \cite{Savchenko2022}. These models classify eight expressions: \textit{Anger, Contempt, Disgust, Fear, Happiness, Neutral, Sadness,} and \textit{Surprise}, achieving 63.03\% accuracy across all 8 categories and 66.34\% when focusing on 7 categories. The models process the same facial crops used for smile detection, providing frame-level expression predictions.

\textbf{Speech Transcription and Text Sentiment Analysis.} Since the collected \emph{Short Interviews} dataset lacks human-annotated speech transcriptions, we employed OpenAI's Whisper model \cite{radford2022robust} to transcribe the interviews for this dataset, extracting segmented utterances along with their timestamps. Additionally, we manually annotated the speaker of each utterance to focus exclusively on participant responses. For the WoZ dataset, we used human-annotated speech transcriptions and sentiment labels already included in the dataset. Once the transcriptions were obtained for both datasets, we applied text sentiment analysis using \emph{pysentimiento} \cite{perez2021pysentimiento}, a Python toolkit for social NLP tasks. The sentiment analysis for Spanish leverages RoBERTuito \cite{perez2022robertuito}, a state-of-the-art language model trained on Spanish social media text. RoBERTuito produces probabilities for each sentiment class (\textit{Positive, Negative,} or \textit{Neutral}), providing the likelihood of each sentiment for every text segment.

\textbf{Temporal Alignment.} To analyze relationships between speech transcriptions and facial behaviors, we align signals using utterances as the temporal unit. For facial signal processing, we downsample video to 10fps to reduce redundancy. The alignment process handles different granularities: text sentiment at utterance-level, smile detection for 16-frame sequences, and facial expressions per-frame. Since facial signals occur more frequently than utterance-level text sentiment, we oversample text sentiment to match facial signal frequency. For a sentiment score of 0.6 with smile events (0.86, 0.87, 0.865), we replicate the sentiment (0.6, 0.6, 0.6) to maintain correspondence and preserve facial dynamics.

\section{Experiments}
\label{sec:experiments}
This section examines facial expressions, text sentiment, and smiles across human-agent and human-human interactions in the WoZ and \emph{Short Interviews} datasets. Our analysis focuses on three aspects: the relationship between human annotations and model predictions, the correspondence between different modalities, and individual user variations. 

\subsection{Comparing Human Annotations and Model Predictions}
\label{study:1}

We studied the alignment between human-labeled annotations and automatic model predictions across two modalities: facial expressions and text sentiment in the WoZ dataset. To quantify the agreement, we generate a co-occurrence matrix that aggregates results for each category and highlights areas of alignment and disagreement.

\begin{figure}[htbp]
 \includegraphics[width=\linewidth]{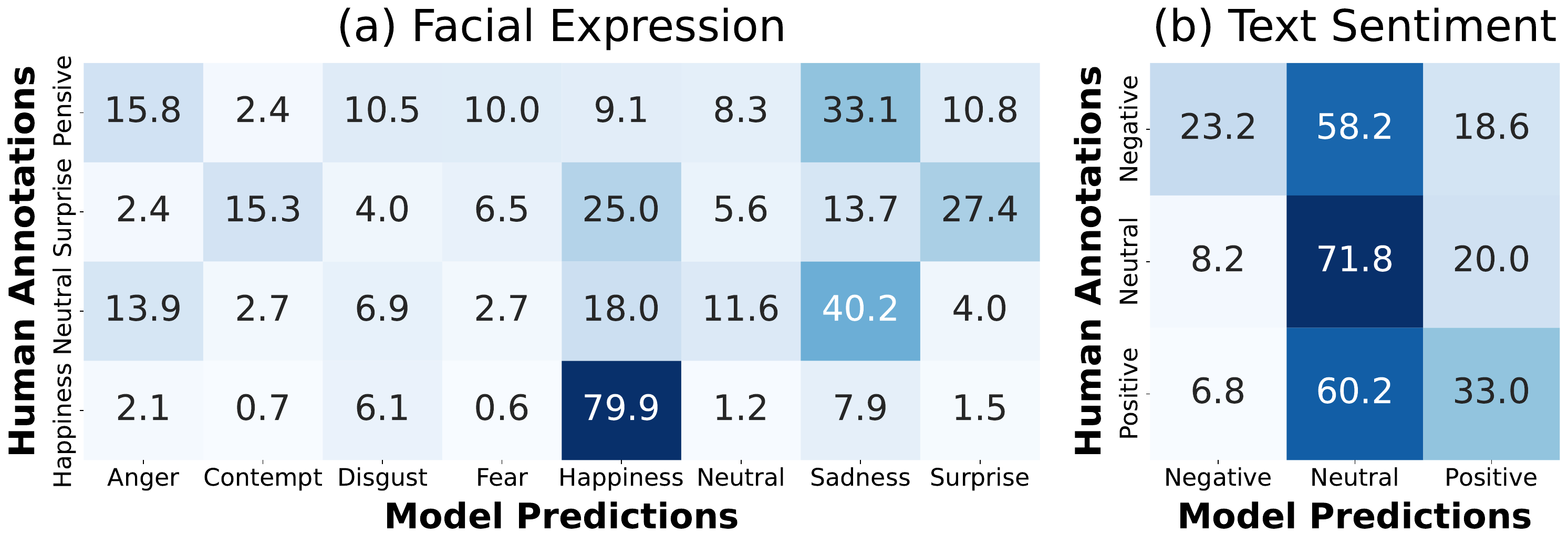}
\caption{Normalized co-occurrence matrices between human annotations and model predictions for (a) facial expressions and (b) text sentiment (zero-shot classification). Values represent percentage distribution for each human annotation category.}
\label{fig:human_and_auto_labels}
\end{figure}

\textbf{Facial Expression.} Comparing human and model annotations reveals that the human annotation categories and automatic model's predefined classes do not share the same set of labels, as shown in Figure~\ref{fig:human_and_auto_labels}a, where annotators used non-standard and subjective categories, making direct comparisons challenging. While \textit{Happiness} shows strong agreement across both annotation approaches, misclassifications arise when human annotators used more subjective categories. For example, human-annotated \textit{Pensive} expressions, which falls outside standard emotion categories, are typically classified by the model as either \textit{Sadness} or \textit{Neutral}. Similarly, when annotators labeled expressions as \textit{Neutral}, the model often interprets these as \textit{Sadness}.

\textbf{Text Sentiment.} The zero-shot performance analysis, presented in Figure~\ref{fig:human_and_auto_labels}b, indicates that the model demonstrates a moderate agreement with human annotations for \textit{Neutral} sentiment. \textit{Positive} texts are frequently classified as \textit{Neutral}, but less frequently confused with \textit{Negative}. Similarly, human-annotated \textit{Neutral} texts are typically classified as either \textit{Neutral} or \textit{Positive. Negative} texts follow a similar pattern, being frequently classified as \textit{Neutral}, but less frequently confused with \textit{Positive}. These findings suggest that while the model effectively distinguishes between  \textit{Positive} sentiment against \textit{Negative}, it struggles to correctly identify \textit{Negative} sentiment, exhibiting a tendency to shift toward neutral or positive predictions. Additionally, the boundary between neutral and polarized sentiments remains ambiguous, reflecting the model’s difficulty in capturing subtle variations in sentiment.

\begin{figure}[htbp]
\includegraphics[width=0.95\linewidth]{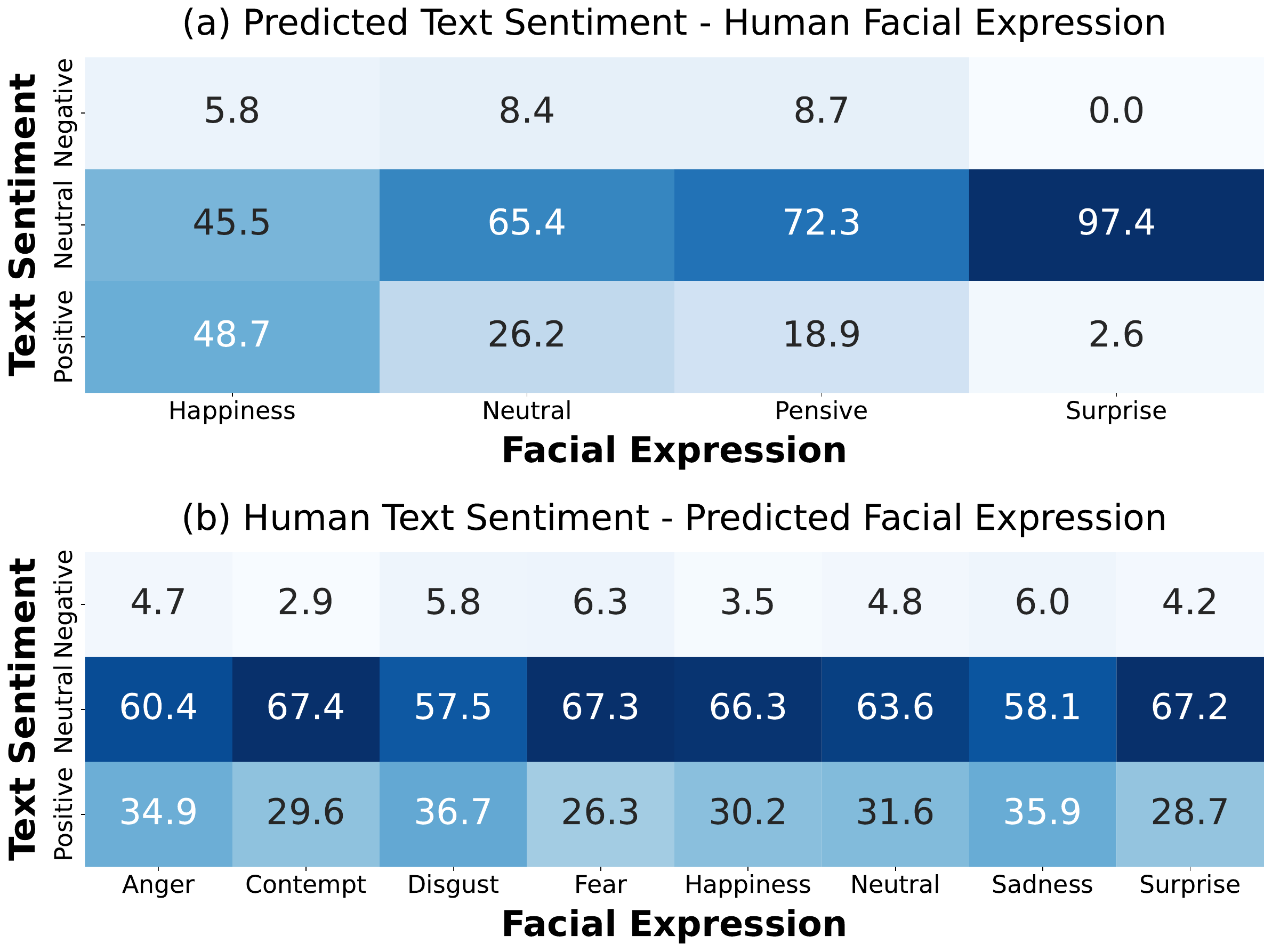}
\caption{Normalized co-occurrence matrices for the WoZ dataset. (a) Model-predicted text sentiment versus human-annotated facial expressions, and (b) human-annotated text sentiment versus model-predicted facial expressions. Values represent percentage distribution for each facial expression category.}
\label{fig:fer_text_sentiment}
\end{figure}

\textbf{Relationship between Text Sentiment and Facial Expression.} We investigate the relationship between facial expressions and speech transcription text sentiment by analyzing their co-occurrence patterns in the WoZ dataset. The modalities were temporally aligned as described in Sec. \ref{sec:methods}. Figure~\ref{fig:fer_text_sentiment} presents two analyses: (a) model-predicted text sentiment versus human-annotated facial expressions, and (b) human-annotated text sentiment versus model-predicted facial expressions. The analysis reveals consistent patterns in text sentiment across both approaches, with \textit{Neutral} sentiment being predominant among all facial expressions (ranging from 45.5\% to 97.4\% for model predictions, and 57.5\% to 67.4\% for human annotations), followed by \textit{Positive} sentiment, while \textit{Negative} sentiment remains consistently low ($<9\%$). Notably, model-predicted facial expressions tend to align with \textit{Neutral} sentiment, occasionally exhibiting \textit{Positive} sentiment. This outcome aligns with the nature of WoZ interactions, which typically involve neutral conversations between users and a virtual agent.

\begin{figure}[h]
\centering
\includegraphics[width=0.6\linewidth]{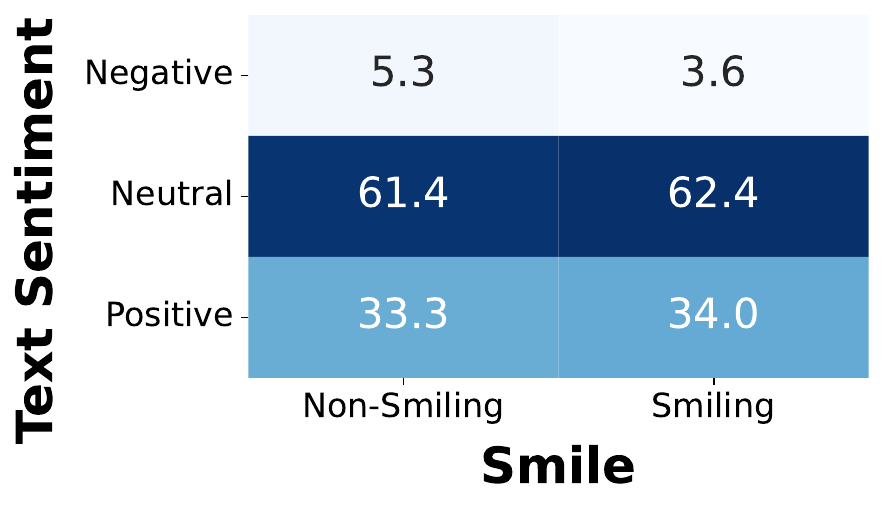}
\caption{Normalized co-occurrence between human-annotated text sentiment and model-predicted smiling states in the WoZ dataset. Values represent percentage distribution for each condition (Smiling and Non-Smiling).}
\label{fig:human_text_smile}
\end{figure}

\textbf{Relationship between Detected Smiles and Human Annotated Text Sentiment}. We examine the relationship between detected smiles and the text sentiment of speech transcriptions. Specifically, we aligned the predominant text sentiment category, extracted from the utterances as described in Sec.~\ref{sec:methods}, with the smile signals occurring during those same utterances. For utterances spanning multiple smile samples, the text sentiment signal was oversampled to match the smile data frequency. Figure~\ref{fig:human_text_smile} presents the distribution of human-annotated text sentiment in relation to automatically detected smiling and non-smiling states. The results show remarkably similar patterns across both conditions: \textit{Neutral} sentiment dominates (61.4\% for non-smiling and 62.4\% for smiling states), followed by \textit{Positive} sentiment (33.3\% and 34.0\% respectively), while \textit{Negative} sentiment remains minimal ($<6\%$). The high co-occurrence of smiles with \textit{Neutral} sentiment suggests that smiles in human-agent interactions might function primarily as social signals of engagement or acknowledgment, aligning with interaction patterns where courtesy smiles serve communicative rather than emotional purposes.

In summary, our analyses reveal both alignments and differences between human annotations and model predictions. The data shows clear agreement in specific cases (e.g., \textit{Happiness} in facial expressions, \textit{Neutral} in text sentiment), while differences emerge in others. In the context of human-agent interactions, our observations suggest that facial signals like smiles and expressions might serve communicative functions beyond emotional role. These findings provide insights into the patterns of human behavior interpretation in human-agent interaction scenarios.

\subsection{Comparing Model Predictions Across different Modalities}

This section compares model predictions across different modalities: facial expressions, text sentiment, and smiles. We examine these relationships across two distinct datasets: WoZ, which captures human-virtual agent interactions, and \emph{Short Interviews}, which contains human-human conversations. We investigate how different behavioral signals align when interpreted by models, and how these patterns vary across different interaction contexts.

\begin{figure}[htbp]
\includegraphics[width=0.95\linewidth]{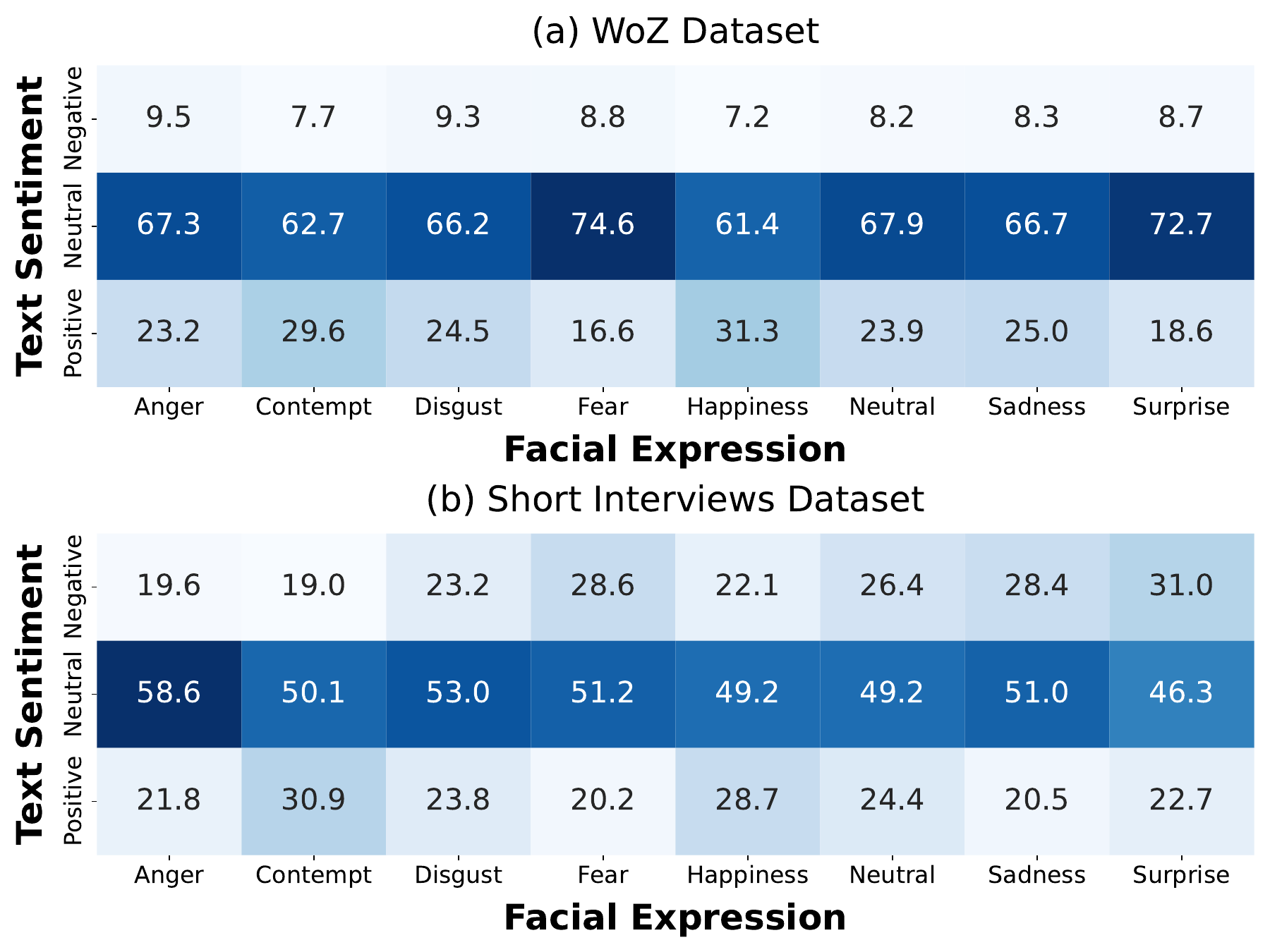}
\caption{Normalized co-occurrence matrices between model-predicted facial expressions and text sentiment. Results shown for (a) WoZ dataset and (b) \emph{Short Interviews} dataset. Values represent percentage distribution for each facial expression category.}
\label{fig:automatic_fer_text_sentiment}
\end{figure}

\textbf{Relationship between Text Sentiment and Facial Expression.} We analyze the patterns between model predictions for facial expressions and text sentiment across two datasets. Figure~\ref{fig:automatic_fer_text_sentiment} shows the co-occurrence patterns between model-predicted facial expressions and text sentiment, where (a) shows results from the WoZ dataset, and (b) from the \emph{Short Interviews} dataset. In the WoZ dataset, similar to our findings in \ref{study:1} \textit{Neutral} sentiment strongly dominates (61.4\% to 74.6\%) across all facial expressions, with \textit{Positive} sentiment following as the second most common prediction, particularly for expressions of \textit{Happiness} (31.3\%) and \textit{Contempt} (29.6\%), while \textit{Negative} sentiment remains consistently low ($<10\%$). The \emph{Short Interviews} dataset, in contrast, presents more balanced predictions, with \textit{Neutral} sentiment ranging from 46.3\% to 58.6\% and notably more balanced polarities. Several facial expressions in this dataset show higher \textit{Negative} than \textit{Positive} sentiment (e.g., \textit{Surprise}: 31.0\% vs 22.7\%, \textit{Sadness}: 28.4\% vs 20.5\%). These differences in sentiment distributions likely reflect the distinct nature of each dataset: WoZ captures human-virtual agent interactions in controlled settings, while the \emph{Short Interviews} dataset contains human-to-human interactions across diverse conversational contexts that may evoke a wider range of emotional responses.

\begin{figure}[htbp]
\includegraphics[width=\linewidth]{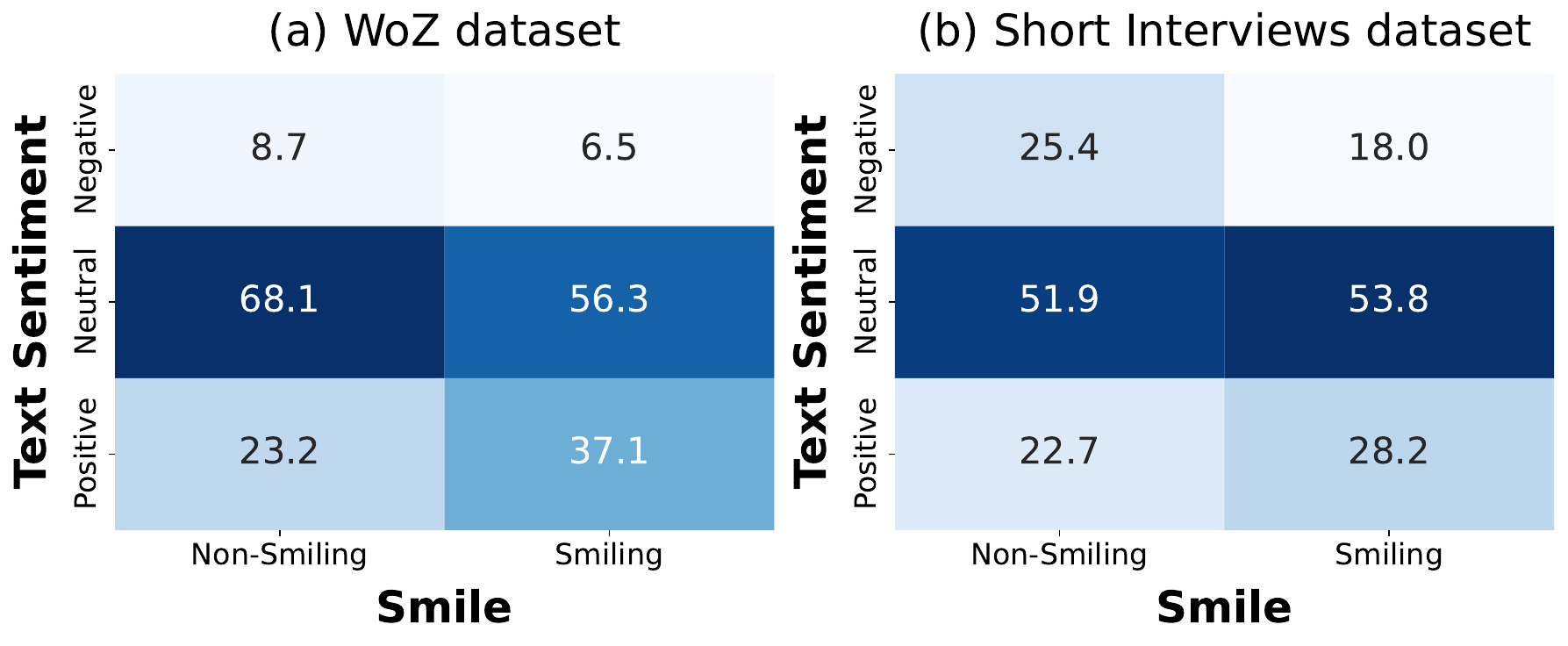}
\caption{Normalized co-occurrence between model-predicted text sentiment and smiling states in (a) WoZ dataset and (b) \emph{Short Interviews} dataset. Values represent percentage distribution for each smiling condition.}
\label{fig:smile_and_sentiment}
\end{figure}

\textbf{Relationship between Smiles and Text Sentiment} We examine the relationship between detected smiles and predicted text sentiment across both datasets. Figure~\ref{fig:smile_and_sentiment} shows the normalized co-occurrence between model-predicted text sentiment and smile states for the WoZ and Short Interviews datasets.
In the WoZ dataset, \textit{Neutral} sentiment dominates across both smiling and non-smiling states, followed by \textit{Positive} sentiment. A notable shift is observed in the distribution of \textit{Neutral} sentiment, which remains more prevalent than other sentiment classes in both conditions. In contrast, the \emph{Short Interviews} dataset reveals a higher presence of \textit{Negative} sentiment during non-smiling events, although \textit{Neutral} sentiment remains dominant across both smile states. Additionally, we observe a low correlation between smile intensity and text sentiment in both datasets. This suggests that smiles may serve broader communicative functions beyond conveying positive emotions, aligning with previous research indicating that smiles play complex roles in human communication and can occur across various emotional contexts \cite{MARTIN2017864}.

Our cross-modal analysis reveals distinct patterns across datasets. While the WoZ dataset shows strong tendencies toward \textit{Neutral} predictions with occasional \textit{Positive} sentiment, the \emph{Short Interviews} dataset exhibits more balanced sentiment distributions. These differences appear consistently across both facial expression and smile analyses, suggesting that the interaction context (human-agent versus human-human) significantly influences model predictions. The weak correlation between smile presence and positive sentiment, observed in both datasets, indicates that behavioral signals in human communication may serve purposes beyond their typically assumed emotional associations.

\subsection{Analyzing individual differences.} 

We analyze the model predictions across different users in WoZ dataset. Our analysis focuses on three aspects: the agreement between model-predicted and human-annotated text sentiment,  the distribution of model-predicted facial expression scores and the distribution of smiling intensity score. We aim to understand how model performance and behavioral predictions vary at the individual level.

\begin{figure}[htbp]
\centering
\includegraphics[width=\linewidth]{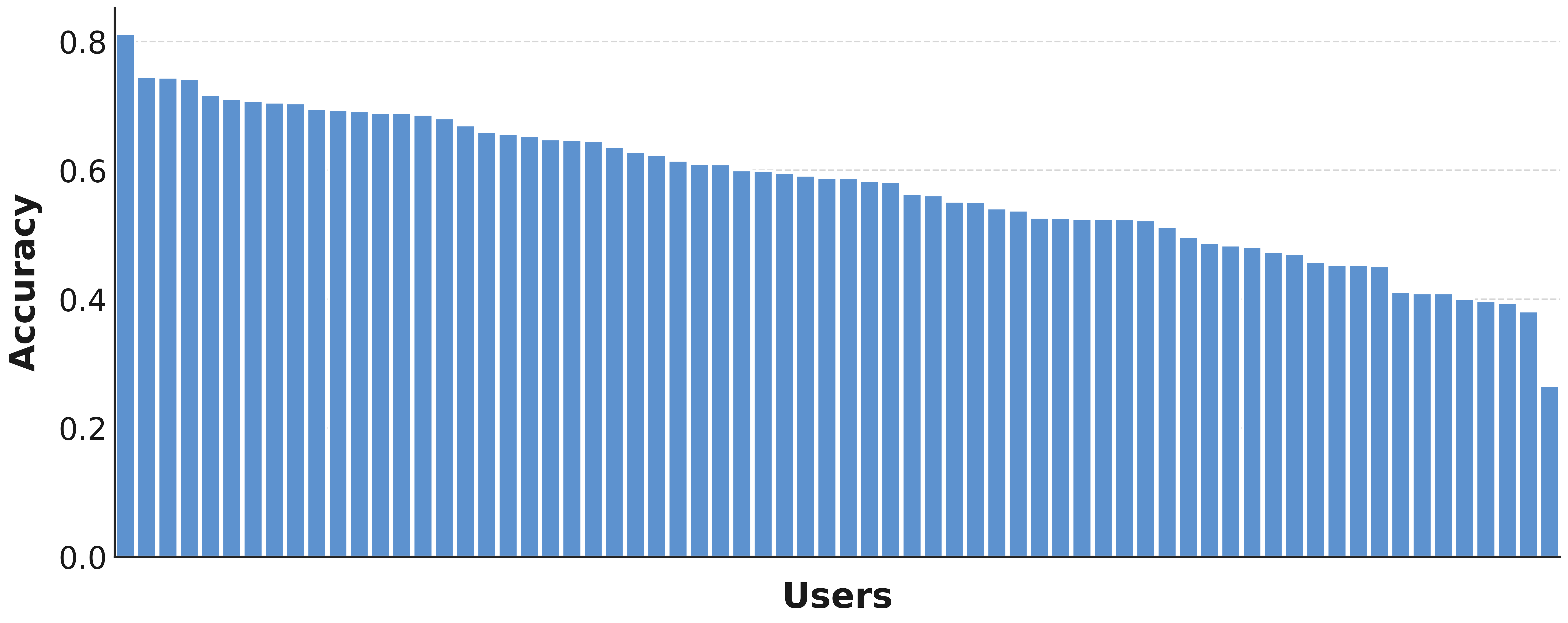}
\caption{Per-user zero-shot accuracy comparison between automatic speech sentiment and human annotations. The figure illustrates that the agreement between automatic sentiment and human annotations varies among users, suggesting model bias towards certain verbal expression patterns.}
\label{fig:accuracy_per_user}
\end{figure}

\textbf{User Variation in Text Sentiment Model Agreement.} The zero-shot text sentiment analysis reveals substantial variation in model performance across users in WoZ dataset, as shown in Figure~\ref{fig:accuracy_per_user}. Agreement between model predictions and human annotations ranges from 0.25 to 0.8, with most users falling between 0.4 and 0.7. This wide range of accuracy scores suggests that the model's ability to capture sentiment varies significantly depending on individual speaking patterns and expression styles. While some users' sentiment expressions align well with model predictions, others show consistent disagreement, indicating that individual verbal expression patterns might not be well represented by the model's general understanding of sentiment.

\begin{figure}[htbp]
\centering
\includegraphics[width=0.95\linewidth]{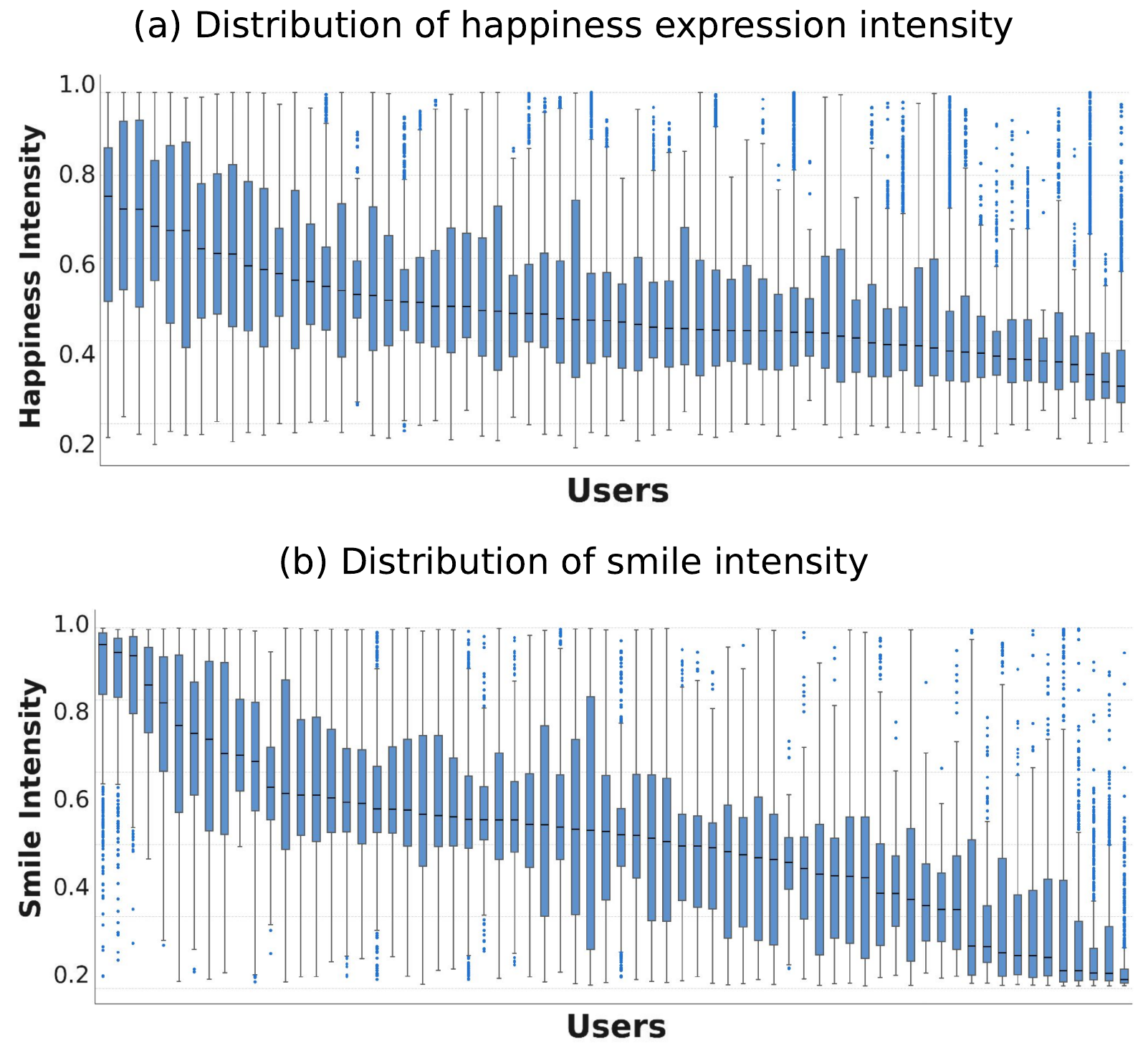}
\caption{Per-user distribution of (a) happiness and (b) smile facial expression intensities shown through boxplots. The figure illustrates that the intensity of facial expressions varies substantially among users.}
\label{fig:expression_boxplots}
\end{figure}

\vspace{1em}
\textbf{User Variation in Facial Emotional Signal.} Figure~\ref{fig:expression_boxplots} presents the distribution of facial emotion scores across users in WoZ dataset for both facial expression of \textit{Happiness} and the intensity of smiles. The boxplots reveal substantial individual differences in how users express themselves facially during interactions. \textit{Happiness} expression intensities (Figure~\ref{fig:expression_boxplots}a) show varying median levels and ranges across users, with some displaying consistently higher intensities (0.6-0.8) and others showing more moderate levels (0.3-0.5). Similarly, smile intensity scores (Figure~\ref{fig:expression_boxplots}b) demonstrate even greater variation, with some users exhibiting frequent high-intensity smiles (0.8-1.0) while others maintain lower intensity ranges (0.2-0.4). The presence of outliers in both distributions suggests that most users occasionally deviate from their typical expression patterns, possibly responding to specific interaction contexts. Similar patterns of individual variation are observed for other facial expressions and across both datasets.

Our analysis reveals two key findings about individual differences. First, the accuracy of text sentiment prediction varies substantially across users, suggesting that models may struggle to capture personal variations in verbal expression. Second, the intensity of facial emotion scores shows distinct patterns for each user, indicating individual styles of non-verbal communication during human-agent interactions. These user-specific patterns in both verbal and non-verbal behavior emphasize the challenges models face in capturing generalizable representations of human expressivity, given the inherent diversity in how individuals communicate and express themselves. This highlights the need to account for individual differences to better understand human behavior.

\section{Discussion}

Our analysis of facial expressions, text sentiment, and smiles in elderly Spanish speakers reveals significant limitations in current affective computing systems. The gap between human annotations and model predictions, particularly evident in facial expression recognition, shows that models often miss the nuanced emotional expressions typical of older adults. These limitations become more pronounced when comparing controlled WoZ interactions with natural conversations in the \emph{Short Interviews} dataset. In the WoZ dataset, where interactions occur between participants and a virtual agent, model outputs across all modalities consistently skew toward positive emotions. This likely reflects the structured and polite nature of virtual agent interactions, which encourage neutral and positive exchanges. In contrast, the \emph{Short Interviews} dataset, featuring more spontaneous human-to-human interactions, exhibits a higher presence of negative sentiment outputs. This distinction emphasizes how the type of interaction context shapes emotional expression and the model predictions.

Our cross-modal analysis reveals important findings about emotional expression in human interactions. The weak correlation between smile intensity and text positive sentiment, combined with frequent smiles during neutral speech, indicates that these expressions serve broader communicative functions. This observation of the multiple purposes of smiles is aligned with previous work \cite{MARTIN2017864}. This understanding is crucial for developing virtual assistants, as these social signals may carry different meanings in human-agent interactions compared to natural conversations. Virtual assistants need to consider not just what an expression is, but what it means in a specific interaction context. 

Individual differences emerge as a critical challenge for current systems. Our findings demonstrate substantial variation in personal expression styles, from varying intensities in facial emotions to distinct patterns in sentiment expression. This consistent variability across both datasets suggests that standard emotion perception approaches may oversimplify human expressivity in this demographic.

These findings have direct implications for developing supportive technologies for older adults. The distinct patterns observed between human-agent and human-human interactions emphasize the importance of context-aware emotion perception. Understanding these nuances is especially important for virtual assistants designed to support elderly well-being, where misinterpreting emotional signals could affect the quality of care. Moreover, the significant individual variations we found suggest the need for more adaptable approaches, especially when working with underrepresented populations like elderly Spanish speakers.

\textbf{Ethical Implications.}
Our study has considered several ethical guidelines. All participants in the \emph{Short Interviews} dataset provided informed consent for the use of their identifiable recordings, and ethical approval was granted by the Ethics Committee of the \emph{Fundació Sant Joan de Déu} healthcare center. To safeguard participant well-being, interviews were carefully designed to elicit only mild emotional responses, avoiding potential distress topics such as personal trauma. This consideration was crucial given the vulnerability of elderly participants, who are more likely to experience feelings of desolation and loneliness. Participants were monitored by an external observer and reminded of their right to withdraw at any time. Although all participants agreed to the use of their data for research purposes, privacy concerns prevent the public sharing of our dataset.

Our research evaluates off-the-shelf affective models using interviews with elderly individuals, with the long-term goal of developing virtual agents and social assistive robots that enhance their well-being. However, such technologies must be applied cautiously to avoid risks like disorientation, deception, or infantilization of users \cite{vercelli2018robots}. While these applications are designed to encourage social interaction, it is important to emphasize that they are meant to supplement, not replace, human interactions. Although our analyses using the \emph{Short Interviews} and WoZ datasets provide valuable insights, we acknowledge their limitations. Emotional responses vary widely due to personal experiences, cultural backgrounds, and personality traits, highlighting the need for personalized models. Moreover, current affective models capture only a limited range of emotions, underscoring the need for further research to achieve more comprehensive and nuanced emotion perception.


\section{Conclusion}
In this work, we evaluated state-of-the-art affective computing models—facial expression recognition, text sentiment analysis, and smile detection—on Spanish-speaking older adults using the WoZ and \emph{Short Interviews} datasets. Our analysis revealed that models struggle to accurately predict the nuanced emotional signals of this demographic. Results highlight the influence of interaction context, showing bias toward neutral/positive emotions in virtual agent interactions, while human-to-human conversations displayed more sentiment variation. The significant individual variability in emotional expression underscores the need for personalized, context-aware models that better account for cultural and contextual factors, ensuring more inclusive emotion perception technologies.

\textbf{Acknowledgements.} This research was partially supported by NHoA project PLEC2021-007868/MICIU/AEI/10.13039/501100011033 and SENTIENT project PID2022-138721NB-I00/MCIN/AEI/10.13039/501100011033. Grants from the Spanish Ministry of Science, the Research National Agency,
and from the European Union "NextGenerationEU/PRTR" and FEDER.

{\small
\bibliographystyle{ieee_fullname}
\bibliography{sample-base}
}

\end{document}